\documentclass[sigconf]{acmart}

\AtBeginDocument{%
  \providecommand\BibTeX{{%
    \normalfont B\kern-0.5em{\scshape i\kern-0.25em b}\kern-0.8em\TeX}}}

\newcommand{\nickname}{GSTalker}
\begin{document}
\pagestyle{empty}

\title{GSTalker: Real-time Audio-Driven Talking Face Generation via Deformable Gaussian Splatting}

%


\author{Bo Chen}
\email{cbforever@sjtu.edu.cn}
 \affiliation{
   \institution{Shanghai Jiaotong University}
   \city{Shanghai}
   \country{China}
   }

\author{Shoukang Hu}
\email{shoukang.hu@ntu.edu.sg}
 \affiliation{
   \institution{Nanyang Technological University}
   \city{Singapore}
   \country{Singapore}
   }

\author{Qi Chen}
\email{cq1073554383@sjtu.edu.cn}
 \affiliation{
   \institution{Shanghai Jiaotong University}
   \city{Shanghai}
   \country{China}
   }
\author{Chenpeng Du}
\email{duchenpeng@sjtu.edu.cn}
 \affiliation{
   \institution{Shanghai Jiaotong University}
   \city{Shanghai}
   \country{China}
   }
\author{Ran Yi}
\email{ranyi@sjtu.edu.cn}
 \affiliation{
   \institution{Shanghai Jiaotong University}
   \city{Shanghai}
   \country{China}
   }
\author{Yanmin Qian}
\email{yanminqian@sjtu.edu.cn}
 \affiliation{
   \institution{Shanghai Jiaotong University}
   \city{Shanghai}
   \country{China}
   }
\author{Xie Chen}
\email{chenxie95@sjtu.edu.cn}
 \affiliation{
   \institution{Shanghai Jiaotong University}
   \city{Shanghai}
   \country{China}
   }




\begin{abstract}
We present \nickname{}, a 3D audio-driven talking face generation model with Gaussian Splatting for both fast training (40 minutes) and real-time rendering (125 FPS) with a 3$\sim$5 minute video for training material, in comparison with previous 2D and 3D NeRF-based modeling frameworks which require hours of training and seconds of rendering per frame. 
Specifically, GSTalker learns an audio-driven Gaussian deformation field to translate and transform 3D Gaussians to synchronize with audio information, in which multi-resolution hashing grid-based tri-plane and temporal smooth module are incorporated to learn accurate deformation for fine-grained facial details.
In addition, a pose-conditioned deformation field is designed to model the stabilized torso. 
To enable efficient optimization of the condition Gaussian deformation field, we initialize 3D Gaussians by learning a coarse static Gaussian representation. 
Extensive experiments in person-specific videos with audio tracks validate that \nickname{} can generate high-fidelity and audio-lips synchronized results with fast training and real-time rendering speed.
\end{abstract}





\maketitle
\section{INTRODUCTION}
Audio-driven talking face generation has received increasing attention due to its promising prospect in various applications, including digital humans, virtual video conferencing, visual dubbing, etc. 
Recent methods~\cite{prajwal2020lip,zhou2020makelttalk,lu2021live,gururani2023space, du2023dae,shen2023difftalk} either learn 2D audio-driven talking faces through direct mapping from audio to face landmarks or 3D audio-driven talking faces with neural radiance field (NeRF) implicit representation. 
However, these methods typically require expensive computational resources for training and rendering, posing potential challenges to their real-world applications. 
For example, although NeRF-based 3D audio-driven talking face models achieve consistent multi-view rendering results, they consume up to 36 hours of training time on a 5-minute Obama video \cite{guo2021ad} to learn a 3D audio-driven talking face model with a rendering speed of 0.08 frames per second (FPS). 


To improve the optimization speed of audio-driven talking face models, a recent line of research~\cite{tang2022real, li2023efficient} focuses on learning efficient 3D representation for 3D audio-driven talking face models. 
For example, instead of learning neural radiance field from coordinate neural networks, RAD-NeRF and ER-NeRF incorporate multi-resolution hashing grid~\cite{muller2022instant} representation with pruning techniques like occupancy grid pruning to improve training and rendering speed.   
Although these methods can accelerate the training and rendering by a large margin, they still require hours of training time and their rendering quality can be improved especially for detailed facial parts like teeth.
A bottleneck hindering their training and rendering efficiency is that the backward mapping-based ray-cast rendering algorithm performs volume rendering on hundreds of sampled points for each pixel.


Recently, point-based 3D Gaussian Splatting (3D-GS)~\cite{kerbl20233d} has been proposed to fast render static~\cite{kerbl20233d} and dynamic scenes~\cite{luiten2023dynamic, yang2023deformable, wu20234d} through splatting a restricted set of discrete 3D Gaussians. 
Motivated by these progresses, we attempt to introduce 3D-GS to 3D audio-driven talking face generation. 
Although this venue is promising, there exist two main challenges we need to tackle: 
1) \textit{How to utilize audio information to drive the 3D talking face and torso in a 3D Gaussian Splatting framework?}
The variety and abstraction of speech signals make speech-driven tasks different from previous static or dynamic scene reconstruction tasks.
A new framework tailored for 3D audio-driven talking face generation should be designed by conditioning audio information in the 3D Gaussians Splatting framework.
2) \textit{How to optimize the efficiency the 3D audio-driven talking face generation model?}
Current 3D-GS methods initialize 3D Gaussians either from a random point cloud or a sparse point cloud produced from Structure-from-Motion (SfM)~\cite{schonberger2016structure, snavely2006photo}.
These two initialization approaches are mainly designed for static scenes with multi-view images, which overlooks the internal structure of taking faces. 
For example, we observe that 3D Gaussians inherited from the random point cloud are difficult to converge in the 3D audio-driven talking face generation task.


To address the above two challenges, we propose a novel framework called GSTalker: which enables rapid training and real-time rendering of audio-driven talking face generation via deformable Gaussian Splatting. 
The detailed strategies are shown as follows. 
1) \textit{Deformable Gaussian Splatting for Audio-Driven Talking Face Generation.} 
Inspired by previous research in deformable neural radiance fields~\cite{park2021nerfies}, we encode a motion-independent static taking face in a canonical space with 3D Gaussians and model a deformable field for position and covariance offsets with audio information or head pose as a condition. 
To learn an accurate deformable mapping from audio to offsets of static 3D Gaussians, we introduce a tri-plane multi-resolution hash encoder to represent Gaussian spatial information and a temporal smooth
module to smooth the audio representation extracted by a pre-trained ASR model~\cite{hsu2021hubert,amodei2016deep,baevski2020wav2vec}.
For each video frame, the tri-plane features and smoothed audio representation are concatenated into a tiny MLP deformable field to predict the position and covariance offsets of the static 3D Gaussians.
With the audio-driven deformable 3D Gaussians, we can fast render the talking face through a tile-based differentiable rasterizer. 
2) \textit{Efficient optimization of 3D audio-driven taking face generation model.}
To achieve efficient optimization, we maintain a static Gaussian initialization from talking face images instead of randomly initiating the 3D Gaussians. 
Specifically, we first use the automatic parsing method \cite{lee2020maskgan} to initialize the head and the torso part respectively.
For the head region, we optimize a set of static 3D Gaussians with all talking face images and use it as the initialization for 3D Gaussians in the head region.
For the torso part, we optimize the 3D Gaussians with a fixed camera stance as the torso part approximates the translation motion in the 2D plane.
Thanks to the rendering speed of the point-based Gaussian Splatting, the above initializations can be finished in about one minute. 
Since the above initialization contains information about talking face images, the deformable field can be quickly optimized to learn audio-dependent offsets of 3D Gaussians for eventual talking face generation. 

Our main contributions are as follows:
\begin{enumerate}
    \item We present GSTalker, a 3D audio-driven talking face generation model with Gaussian Splatting for both fast training (40 minutes) and real-time rendering (125 FPS), in which an audio-driven deformable field together with a multi-resolution hashing grid-based tri-plane and a temporal smooth module is incorporated to learn fine-grained facial details.
    \item To enable efficient optimization of audio-driven talking face generation, we learn static initializations of 3D Gaussians for head and torso regions from talking face images.
    \item Extensive experiments in person-specific videos with audio tracks validate that GSTalker can generate high-fidelity and audio-lips synchronized results with fast training and real-time rendering speed.
\end{enumerate}
\section{RELATED WORK}
\subsection{Audio-Driven Talking Face Generation}
Audio-driven talking face generation aims to synthesize synchronized speaking videos of a specified person given the input audio. 
Recently there have been active and consistent research efforts on generating high-fidelity images and achieving better lip-audio synchronization,
the methods they have proposed can be categorized into 2D-based method \cite{thies2020neural,lu2021live,gururani2023space,du2023dae, stypulkowski2023diffused} and 3D NeRF-based method \cite{guo2021ad,liu2022semantic, ye2023geneface,tang2022real}. 
Wav2lip \cite{prajwal2020lip} proposes to use a pre-trained lip-sync expert model as the discriminator for lip-syncing a talking face video and performs adversarial training. 
Neural voice puppetry (NVP) \cite{thies2020neural} leverages 3DMM coefficients and neural textures to synthesize mouth movement. 
However, these two methods can only synthesize the mouth but fail to control the head pose. 
SPACE \cite{gururani2023space} proposes a latent representation based on a pre-trained face-vid2vid model to achieve one-shot video synthesis but still struggles for multi-view consistency. 
Recently, the Diffusion Denoising Probabilistic Model (DDPM) \cite{ho2020denoising} exhibits unique advantages in talking face generation. 
In DAE-Talker \cite{du2023dae}, a diffusion autoencoder (DAE) is first pre-trained on face images and attains a well-defined latent space, and then the model is trained to predict the latent representations of the DAE from speech. Diffused Head \cite{stypulkowski2023diffused} proposes an auto-regressive frames prediction manner with diffusion models. 
Diffusion-based methods have achieved impressive image quality but slow inference.

NeRF \cite{mildenhall2021nerf} is another promising direction and has been widely explored in audio-driven talking face generation. 
AD-NeRF \cite{guo2021ad} maps the audio features to dynamic neural radiance fields for portrait rendering. 
Geneface \cite{ye2023geneface} improves the generalizability of the  NeRF-based methods by introducing a pre-trained generative audio-to-motion model. 
RAD-NeRF \cite{tang2022real} firstly introduces multi-resolution hash grids to encode spatial and phonetic information separately, which greatly accelerates NeRF-based method training and inference. 
ER-NeRF \cite{li2023efficient} leverages tri-plane hash representation to reduce hash collision and thus achieve fast convergence. and a cross-modal fusion mechanism has been proposed to enhance lip-speech synchronization.
Although NeRF-based methods can achieve photorealistic image quality and multi-view consistency, their training and inference are still computationally expensive because of the ray-cast operation and volume rendering computation in NeRF.
\subsection{Efficient Neural Representation}
In some scenarios, the generation of talking-head videos needs to operate in real-time.
In light of the intensive training cost and slow inference lied in vanilla NeRF \cite{mildenhall2021nerf} approaches, various hybrid explicit-implicit representations \cite{sun2022direct, chen2022tensorf, muller2022instant} are proposed to accelerate neural rendering. 
Instant-NGP \cite{muller2022instant} adopts a multi-resolution hash table to replace parts of the computationally expensive fully connected layers, significantly improving rendering efficiency without loss of computational precision.
Recently 3D Gaussian Splatting (3D-GS) \cite{kerbl20233d} achieves the state-of-the-art visual quality and real-time rendering for novel-view synthesis. 
Specifically, 3D-GS adopts 3D Gaussians as a discrete representation of the scene and allows parameter optimization based on differentiable anisotropic splatting. 
Compared to the methods based on volume rendering of implicit neural radiance fields, 3D-GS reduces unnecessary spatial computation and the fast visibility-aware rendering algorithm achieves parallel optimization across pixels. 
Since 3D-GS is only suitable for multi-view reconstruction of static scenes,
many methods \cite{luiten2023dynamic, yang2023deformable, wu20234d} have been proposed to achieve dynamic scene reconstruction based on 3D-GS. 
Dynamic3DGS \cite{luiten2023dynamic} creates positions and properties for 3D Gaussians at each timestamp to achieve dynamic scene reconstruction. 
4D-GS \cite{wu20234d} proposes time-conditioned deformation fields to predict the deformation and movement of Gaussian. 
Recently, many approaches \cite{zielonka2023drivable, qian20233dgs, hu2023gauhuman, xu2023gaussian, hu2023gaussianavatar, lei2023gart} have also tried to build dynamic 3D human models based on 3D-GS and parametric human model SMPL \cite{loper2023smpl}.
Our work first introduces 3D-GS to audio-driven talking face generation, enabling faster convergence and high-fidelity video generation.
\section{method}
\begin{figure*}[h]
    \centering
    \includegraphics[width=1.0\linewidth]{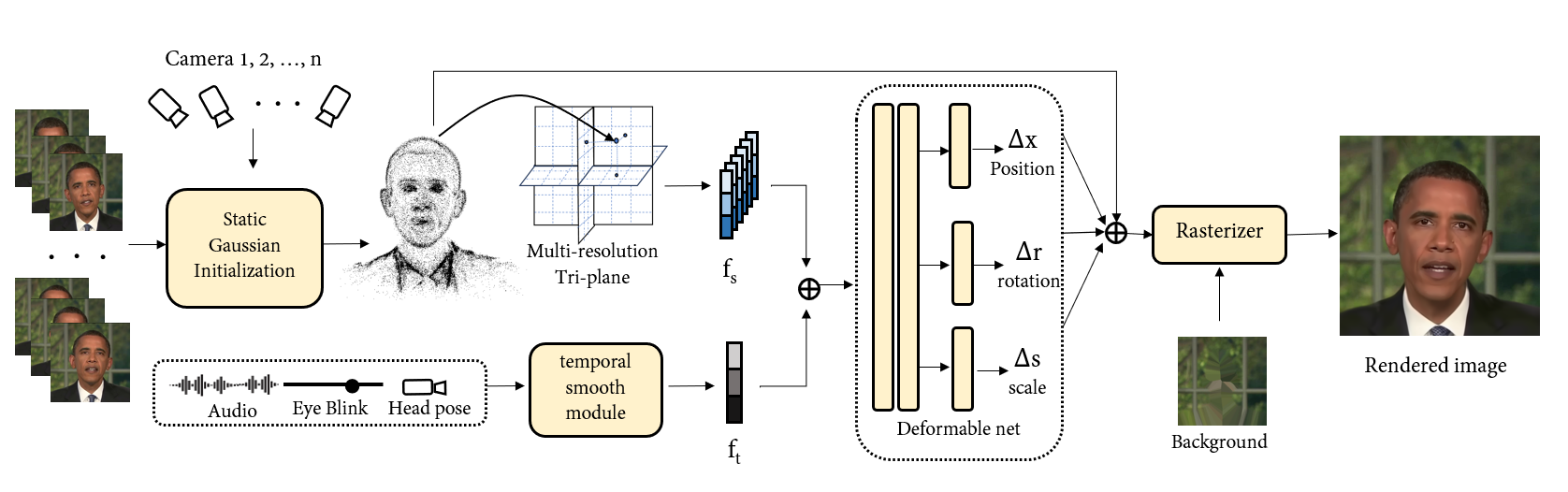}
    \caption{\textbf{Overview of GSTalker.} We model the whole talking face in 3D Gaussians and deformation fields. Firstly, a static initialization stage is used to optimize the coarse static Gaussian of the face from a random point cloud. 
    Then an audio-conditioned deformation field fuses spatial features from a multi-resolution tri-plane hash grid with audio features from an audio encoder to predict the position and shape changes of the 3D Gaussians. Given a camera pose, the deformed Gaussians achieve real-time rendering using the differentiable rasterizer. For the torso part, a similar pose-condition deformation field is adopted to drive the stabilizing motion of the torso.}
    \label{fig:pipeline}
\end{figure*}

 Similar to previous NeRF-based methods \cite{guo2021ad, tang2022real, li2023efficient}, 
Given a 3-5 minute video of a specified portrait speaking in front of a fixed camera as training material, we aim to synthesize a talking face video of the portrait with a new speech.
We use the camera pose estimated by the face-tracker \cite{thies2016face2face} to simulate the head motion.
For audio signal processing, we use a pre-trained ASR model \cite{amodei2016deep,baevski2020wav2vec} for feature extraction.
Since the motion of the head and torso parts is inconsistent, we use the Semantic Parsing Method \cite{lee2020maskgan} to segment the head, torso, and background parts. 
In the following part, we first review the preliminaries of 3D Gaussian Splatting \cite{kerbl20233d} in Sec. \ref{Pre}. 
Then we elaborate on the proposed real-time audio talking face generation framework via deformable Gaussian Splatting in Sec. \ref{gstalker}, followed by the details of the efficient optimization as presented in Sec. \ref{train}.

\subsection{Preliminaries} \label{Pre}
\textbf{3D Gaussian Splatting}.
In 3D Gaussian Splatting \cite{kerbl20233d}, a static scene can be represented by a 3D Gaussian set. Each 3D Gaussian consists of a series of optimizable parameters, including a mean position $\mu$, positive semi-definite covariance matrices $\Sigma$, e.t.,
\begin{equation}\label{eq:1}
G(x)=e^{-\frac{1}{2}(x-\mu)^T\Sigma^{-1}(x-\mu)}
\end{equation}

To facilitate optimization, $\Sigma$ is usually decomposed into a scaling matrix $S$ and rotation matrix $R$:
\begin{equation}\label{eq:2}
\Sigma=RSS^TR^T
\end{equation}
and each matrix can be converted into two optimizable vectors: a 3D vector $s\in R^3$ for scaling and a quaternion $q\in R^4$ for rotation.

Then we transform the 3D Gaussian from world space to image space. As shown in \cite{zwicker2001ewa}, given a viewing transformation $W$ and a affine approximation Jacobian $J$ of the projective transformation, the covariance matrix $\Sigma'$ in ray coordinates is computed by following:
\begin{equation}\label{eq:3}
\Sigma'=JW\Sigma W^TJ^T    
\end{equation}

For each pixel on the image, the color $C$ is computed by blending N ordered points overlapping the pixel:
\begin{equation}\label{eq:4}
C=\sum_{i\in N}c_i\alpha_i\prod_{j=1}^{i-1}(1-\alpha_j)
\end{equation}
where $c_i$ is represented by spherical harmonics (SH) coefficients and $\alpha_i$ is the corresponding learned opacity.

\textbf{}

\subsection{Deformable Gaussian Splatting for Talking Face} \label{gstalker}
Previously, the NeRF-based method \cite{guo2021ad, tang2022real, li2023efficient} directly condition implicit function to model the topological changes of the talking face: $F: f,g,d \rightarrow c,\sigma$, where $f,g,d$ are the spatial feature, audio feature, and view direction respectively. 
Nevertheless, encoding 3D Gaussian directly for each audio feature is inefficient in expressing dynamic scenes.
Noting that only a small portion of the facial regions in the talking face have significant topological deformations, such as the mouth and eyes, it is intuitively possible to maintain a static 3D Gaussian $G$ in a canonical space to characterize the physical geometry of the head and use a deformation field $F$ to learn the audio-driven deformation information about 3D Gaussian.
Since each 3D Gaussian has a different level of correlation with the audio features, for example, the points in the mouth region are more closely affected by audio and have more frequent movement, we take the spatial features of the different 3D Gaussian and the audio features $A$ as inputs to the deformation field to predict the deformed 3D Gaussian $G'$. Then the deformed 3D Gaussian renders the
talking face image $I$ through a tile-based differentiable rasterizer. The whole process can be formulated as follows:
\[I = Splatting(G'|G,F,A)\tag{5}\]
\textbf{Multi-resolution Hashing Grid-based Tri-plane}.
Multiresolution Tri-plane Hash encoder utilizes a structured mesh to store latent features instead of a partially computationally expensive fully connected layer, and the multiresolution feature fusion well considers the motion similarity of neighboring Gaussian points.
The Tri-plane hash method simultaneously reduces unnecessary computations in Gaussian sparse regions and hash conflicts in Gaussian dense regions.
Specifically, as shown in Figure \ref{fig:pipeline}, we first project the positions $\textbf{x}=(x,y,z)\in R^3$ of each 3D Gaussian into a multi-resolution hash tri-plane grid, and the projected coordinate will be interpolated over different resolution levels of 2D hash grid to obtain high dimensional optimizable spatial features $f_H$. 

To reduce the noise between voice frames and enhance the temporal correlation between generated video frames, the audio HuBERT encoder~\cite{hsu2021hubert} successively utilizes 1D convolutional networks and self-attention modules to temporally smooth the audio features extracted from the pre-trained ASR models into feature $f_a$. To control the blinking action, as in RAD-NeRF \cite{tang2022real}, we compute the ratio of the open eye area to the whole image area as our eye feature $f_e$.
These features will be concatenated and fed into a tiny MLP decoder to predict the positional movement $\Delta \textbf{x}$, rotation change $\Delta q$, and scaling change $\Delta s$ of 3D Gaussians:
\begin{equation}
    \Delta \textbf{x},\Delta q, \Delta s = MLP(f_H, f_a, f_e).
\end{equation}
Then the deformed 3D Gaussians is formulated as $G(\textbf{x}+\Delta x, q+\Delta q, s+\Delta s, SH, \alpha)$. 
In our framework, we assume the audio information has a negligible effect on the deformation of spherical harmonic coefficients (SH) and opacity $\alpha$.
We keep the SH coefficients and opacity $\alpha$ of deformable 3D Gaussians unchanged during the audio-driven deformation. 
This design allows us to employ a lighter network for the deformation network and accelerates the training process.


Since the torso usually orients the camera frontally and its movement is a tiny translation motion, we adopt a similar setting as AD-NeRF \cite{guo2021ad}: using head pose $\boldsymbol{P}=(\boldsymbol{R},t)$ rather than audio as the condition to drive the torso's movement, where $\boldsymbol{R}, t$ are the rotation and translation matrix respectively.
Specifically, we train a  Gaussian deformation field with head pose $\boldsymbol{P}$ as the condition. To better inscribe the details, before we feed the head pose $\boldsymbol{P}$ into the deformation field, we use positional encoding \cite{mildenhall2021nerf} $\gamma(p) = (sin(2^i\pi p), cos(2^i\pi p))_{i=0}^{L-1}$ to map it to a higher-dimensional feature $f_p$,
where L=6 in our experiments. The pose features $f_p$ and the spatial feature $f_H$ encoded from the hash grid are then concatenated together to predict the motion and deformation of 3D Gaussians:
\begin{equation}
    \Delta \textbf{x},\Delta q, \Delta s = MLP(f_H, f_p).
\end{equation}
Since the torso's movement involves only one viewpoint, the deformed 3D Gaussians $G(\textbf{x}+\Delta x, q+\Delta q, s+\Delta s, SH, \alpha)$ is projected under a fixed view $\boldsymbol{\Pi_0}=(\boldsymbol{R_0},t_0)$ for differentiable splatting.


\subsection{Efficient Optimization of 3D Audio-Driven Talking Face Generation}
\label{train}
Compared with the disordered and random Gaussian initialization, the Gaussian initialization that can preliminarily characterize the geometry makes it easier to obtain a more convenient deformation process with specific physical meaning.
Unlike the current 3DGS-based avatar which a 3D parametric model is used geometrically before Gaussian initialization, we get a coarse static 3D Gaussian representation in canonical space without adding a neural network.

\begin{table*}[t]
\begin{tabular*}{\textwidth}{@{\extracolsep{\fill}}lcccccccc}
  \toprule
  Methods         &PSNR$\uparrow$    &LPIPS$\downarrow$  &LMD$\downarrow$  &Sync$\uparrow$  &AUE$\downarrow$   &Trainging time  &Inference FPS \\ 
  Ground Truth    & $\infty$  &0      &0    &7.491 &0     &-               &-\\
  \hline
  MakeItTalk\cite{zhou2020makelttalk} &-    &-  &5.481  &5.112  &1.519  &-  &12 \\
  Wav2lip\cite{prajwal2020lip}  &-   &-  &3.718  &\textbf{6.892}  &1.218  &-  &15 \\
  AD-NeRF\cite{guo2021ad}   &31.39  &0.0857 &2.917  &5.426  &0.991  &36h    &0.08 \\
  RAD-NeRF\cite{tang2022real}   &34.00 &0.0387  &2.696  &6.664  &0.882  & 7h     &40  \\
  ER-NeRF\cite{li2023efficient}   &33.10 &0.0291  &2.740  &5.708  &\textbf{0.854}  &2h    &34 \\
  \hline
  GSTalker  &\textbf{34.65} &\textbf{0.0151} &\textbf{2.695} &5.775  &0.862  &\textbf{40min}  &\textbf{125} \\
  \bottomrule
\end{tabular*}
\bigskip\centering
\caption{\textbf{The quantitative comparison under the self-driven setting.} We show the best results with \textbf{bold}. MakeItTalk is unable to specify head posture, and Wav2lip can see the mouth shape of the ground truth, so we use other voices as audio input. For a fair comparison, we do not compare the PSNR and LPIPS of these two methods. We only compare the training time of person-specific methods.}\
\label{t1}
\end{table*}

\begin{figure*}[!h]
	\begin{minipage}{0.02\linewidth}
	\vspace{-30pt}
	\centerline{\rotatebox{90}{\textcolor{red}{sho}p}}
	\vspace{54pt}
	\centerline{\rotatebox{90}{\textcolor{red}{ba}ck}}
	\vspace{54pt}
	\centerline{\rotatebox{90}{\textcolor{red}{the}}}
\end{minipage}
	\begin{minipage}{0.130\linewidth}
	\vspace{3pt}
	\centerline{\includegraphics[width=\textwidth]{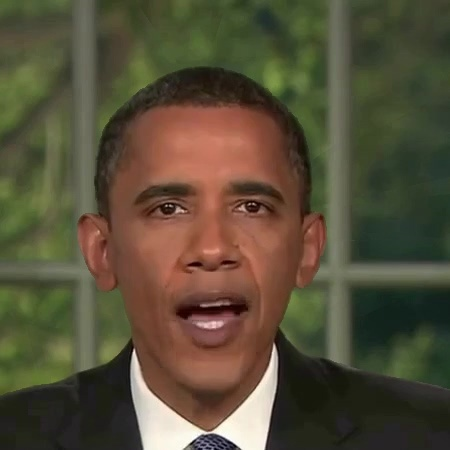}}
	\vspace{3pt}
	\centerline{\includegraphics[width=\textwidth]{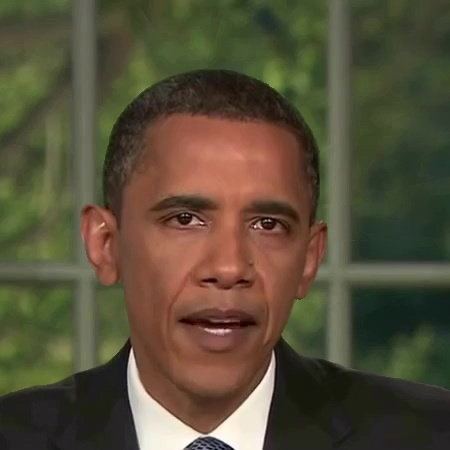}}
	\vspace{3pt}
	\centerline{\includegraphics[width=\textwidth]{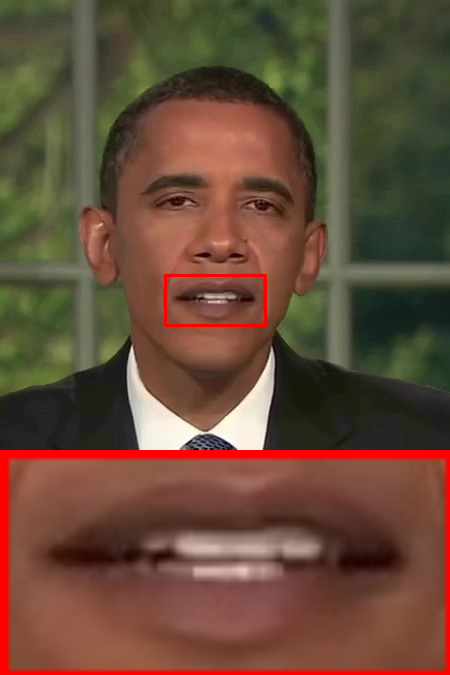}}
	\vspace{3pt}
	\centerline{Ground-Truth}
	\end{minipage}
 	\begin{minipage}{0.130\linewidth}
	\vspace{3pt}
	\centerline{\includegraphics[width=\textwidth]{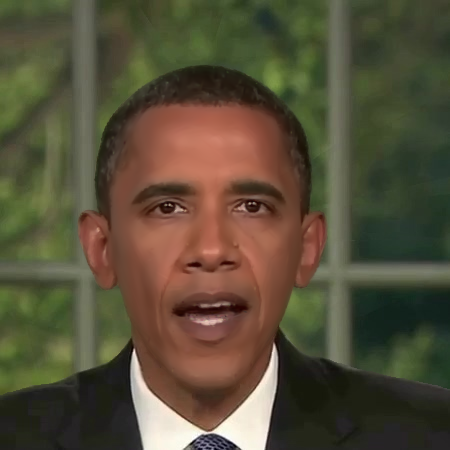}}
	\vspace{3pt}
	\centerline{\includegraphics[width=\textwidth]{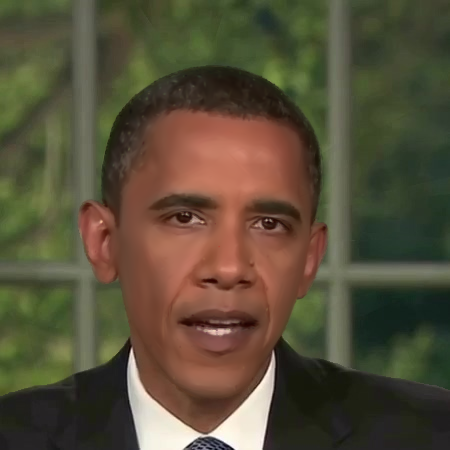}}
	\vspace{3pt}
	\centerline{\includegraphics[width=\textwidth]{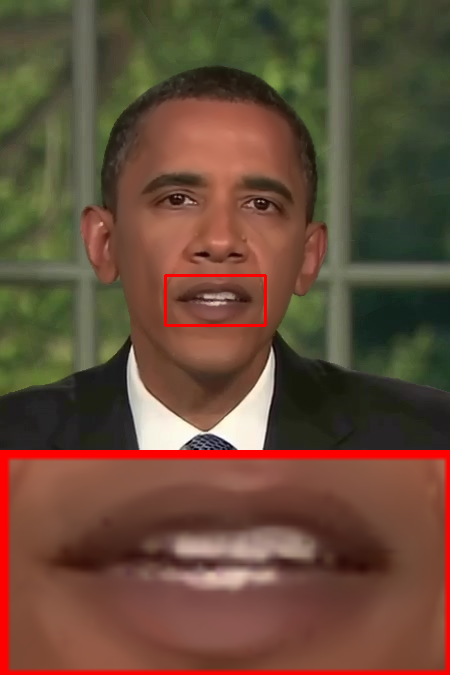}}
	\vspace{3pt}
	\centerline{GSTalker}
\end{minipage}
	\begin{minipage}{0.130\linewidth}
	\vspace{3pt}
	\centerline{\includegraphics[width=\textwidth]{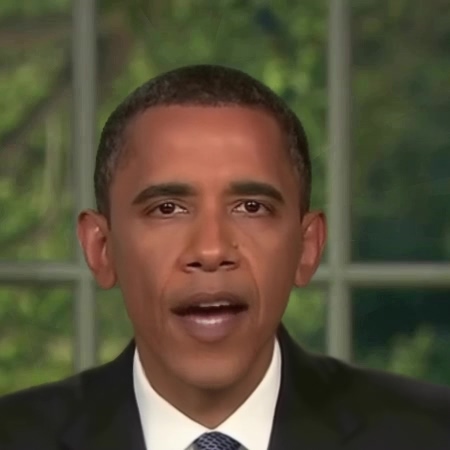}}
        \vspace{3pt}
        \centerline{\includegraphics[width=\textwidth]{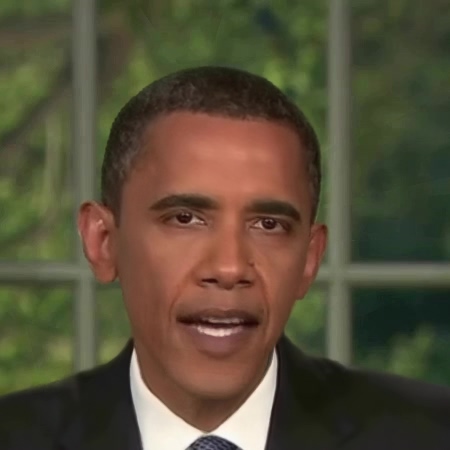}}
        \vspace{3pt}
        \centerline{\includegraphics[width=\textwidth]{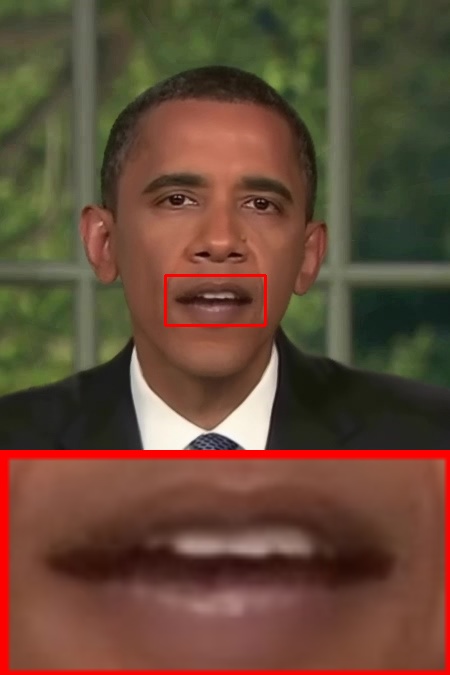}}
	\vspace{3pt}
	\centerline{ER-NeRF}
\end{minipage}
	\begin{minipage}{0.130\linewidth}
	\vspace{3pt}
	\centerline{\includegraphics[width=\textwidth]{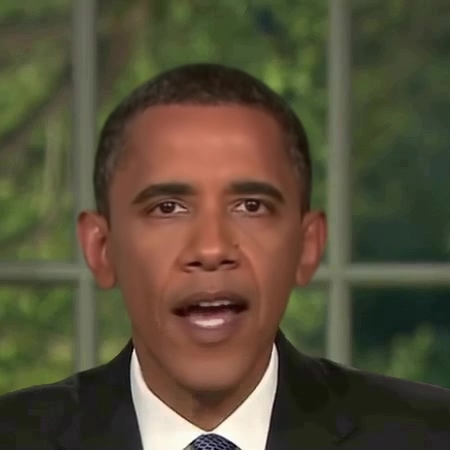}}
	\vspace{3pt}
	\centerline{\includegraphics[width=\textwidth]{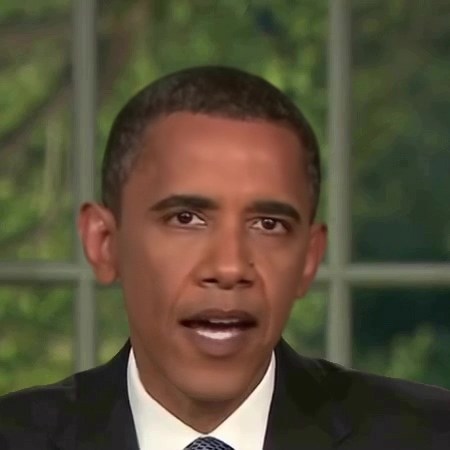}}
	\vspace{3pt}
	\centerline{\includegraphics[width=\textwidth]{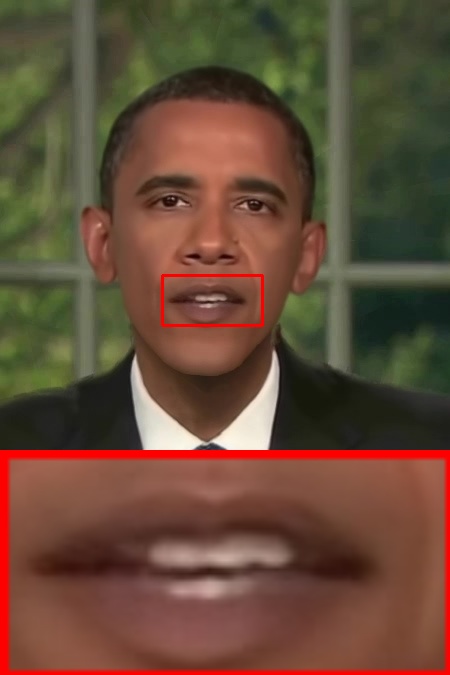}}
	\vspace{3pt}
	\centerline{RAD-NeRF}
\end{minipage}
	\begin{minipage}{0.130\linewidth}
	\vspace{3pt}
	\centerline{\includegraphics[width=\textwidth]{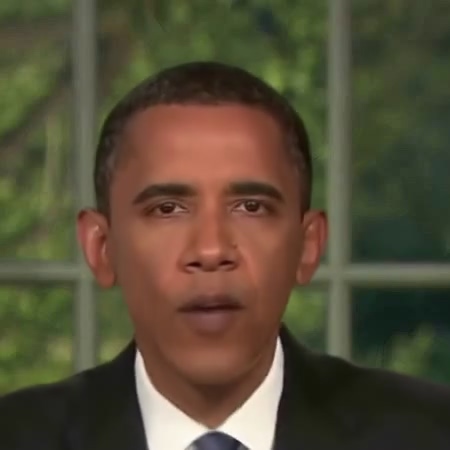}}
        \vspace{3pt}
        \centerline{\includegraphics[width=\textwidth]{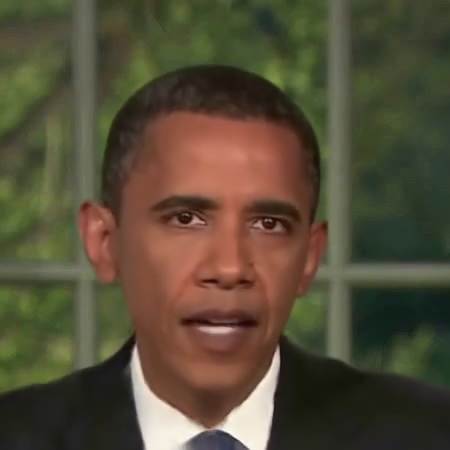}}
        \vspace{3pt}
        \centerline{\includegraphics[width=\textwidth]{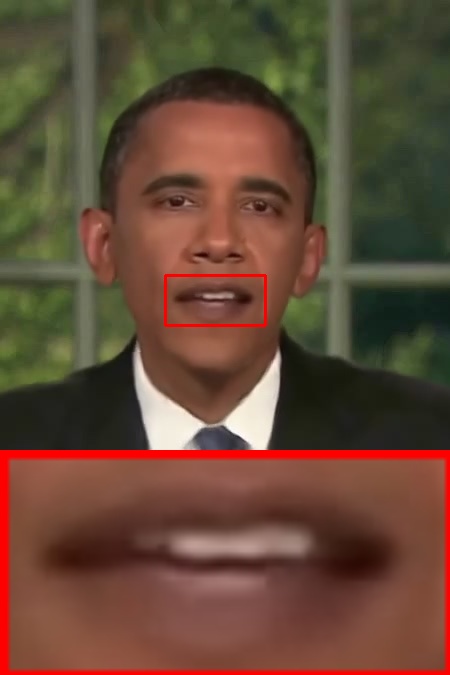}}
	\vspace{3pt}
	\centerline{AD-NeRF}
\end{minipage}
 	\begin{minipage}{0.130\linewidth}
	\vspace{3pt}
	\centerline{\includegraphics[width=\textwidth]{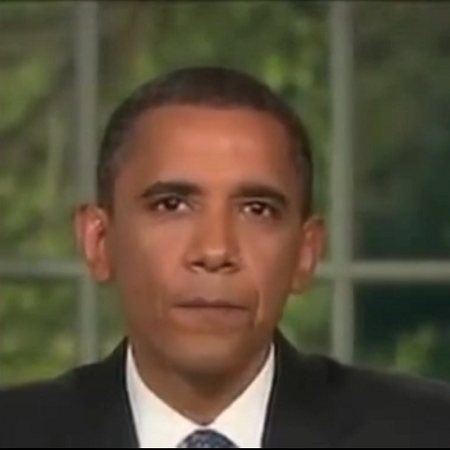}}
	\vspace{3pt}
	\centerline{\includegraphics[width=\textwidth]{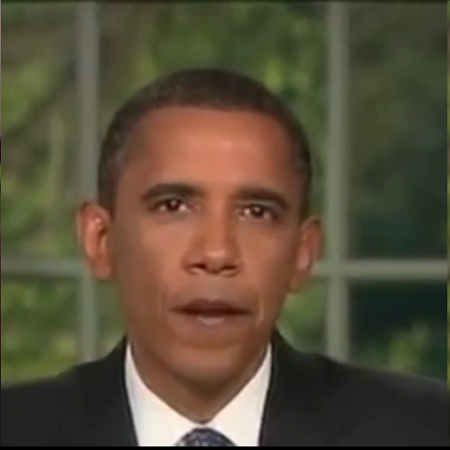}}
	\vspace{3pt}
	\centerline{\includegraphics[width=\textwidth]{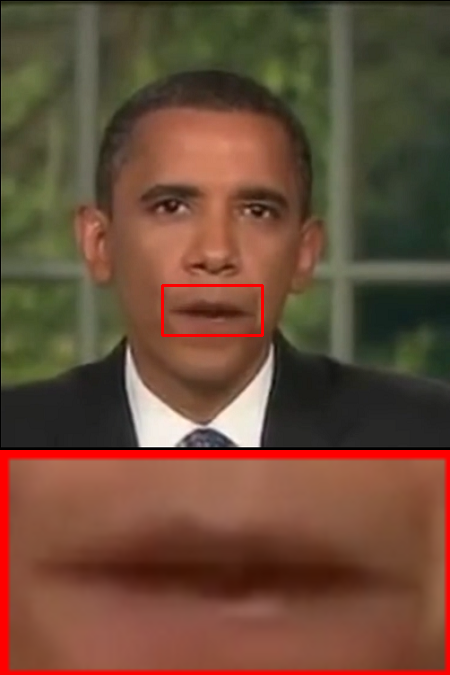}}
	\vspace{3pt}
	\centerline{MaketItTalk}
\end{minipage}
	\begin{minipage}{0.130\linewidth}
	\vspace{5pt}
	\centerline{\includegraphics[width=\textwidth]{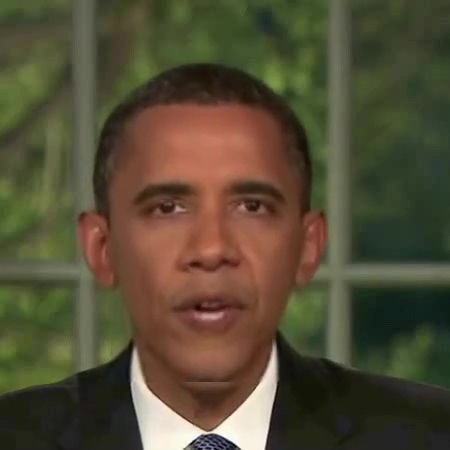}}
	\vspace{3pt}
	\centerline{\includegraphics[width=\textwidth]{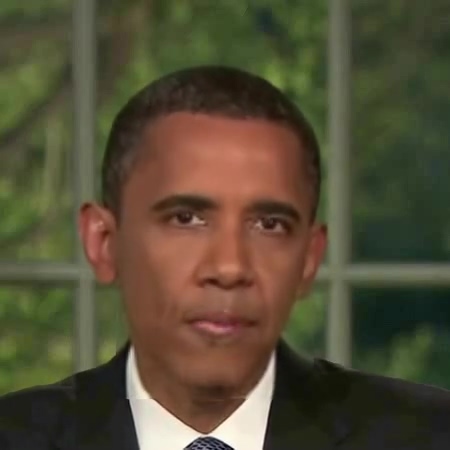}}
	\vspace{3pt}
	\centerline{\includegraphics[width=\textwidth]{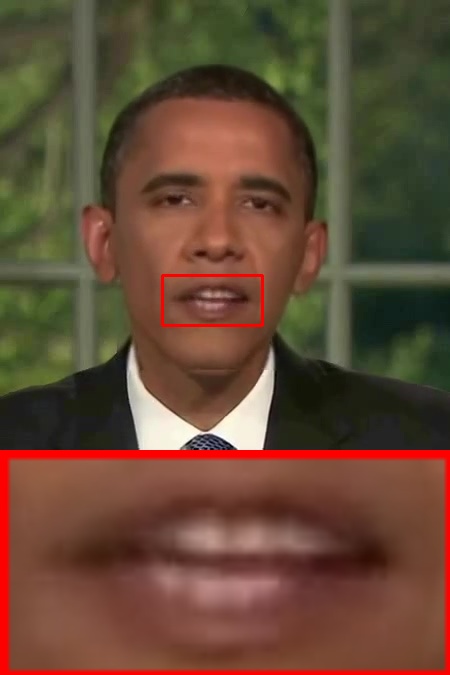}}
	\vspace{3pt}
	\centerline{Wav2Lip}
\end{minipage}
\caption{\textbf{The qualitative comparison with 2D-bsed and 3D NeRF-based Mehods.} We represent the image results generated by our method and all the baselines on the AD-NeRF\cite{guo2021ad} dataset.}
\label{f2}
\end{figure*}

\textbf{Static Gaussian initialization.}
From the perspective of the camera space, the movement of the whole head can be divided into rigid movements related to the head posture, including the translation and rotation of the head and the expression transformation related to facial expressions, while speech is only related to facial expressions including lips and is not correlated with head posture.
Therefore, we explicitly utilize the camera transformation $\Pi$ to simulate the head movement $P$ instead of the model transformation which adopts the additional pose control parameters that need to be learned. On the other hand, the rigid motion of the head does not change the geometry of the head, so this transformation method does not affect the learning of initializing the head geometry.
After that, we regarded the whole process of 3D Gaussian initialization as a static scene reconstruction process, ignoring the movement part of the entire facial expression related to speech, and focusing on learning the static facial structure and skin texture.
Specifically, for each iteration i, the 3D Gaussians is formulated as $G(i) = \{x_i, s_i, q_i, \alpha_i, c_i\}$, we randomly select an image $I_{gt}(A_t, P_t)$ from the monocular video: $\{I(A_1, P_1), ..., I(A_n, P_n)\}$ as the ground truth, where $A$ represents the speech feature and $P$ represents the head pose, then we render the image $I(G(i), R_t, T_t, K)$, where $K$ is the intrinsic matrix and $\{R_t, T_t\}$ is the camera pose which is the inverse of head poses $P_t$. By continuously calculating the loss between $I$ and $I_{gt}$, we obtain an initialized 3D Gaussian representation $G(n) = \{x_n, s_n, q_n, \alpha_n, c_n\}$of the head after n iterations. Our experiments reveal that the initialization of 3D Gaussians has a huge influence on the convergence of the 3D audio-driven talking face generation. 
As mentioned in \cite{guo2021ad}, the camera pose is computed from the head pose and the movement of the head part is not consistent with the movement of the torso part, so we need to divide the portrait image into the head region and torso region at the beginning. 
Since the torso usually orients the camera frontally and its movement is a tiny translation motion, we set a fixed camera pose for rendering and the rest of the optimization process is similar to the head region.
For convenience, we randomly initialize 10,000 3D Gaussians in the space of a cube, then we only need tens of seconds to complete the entire static Gaussian initialization process because the whole process of Gaussian splatting is completely differentiable. 
\textbf{Adaptive Density Control Strategy}.
In addition to optimizing the learnable parameters of 3D Gaussian itself, we also follow the prune, clone, and split strategies in 3D-GS to adaptively control the amount of 3D Gaussians.
Since the clone and split operations are based on the position gradients and magnitude of the scaling matrix, for some regions of frequent movement such as the eyes and mouth, it will gather more and smaller 3D Gaussian. For the less moving part, there will be no more redundant Gaussian because the losses converge faster.
The adaptive density control strategy is more beneficial to more refined and accurate control of the moving part in the later conditional Gaussian deformation field without introducing redundant computations.

\textbf{Training Objectives.}
We use the L1 regularization term $L_{color}$ for each pixel’s color between the rendered image $I$ and the ground truth image $I_{gt}$ which is also used in 3D-GS.
To facilitate a better blend between the rendered head region, torso region, and the background, we further use the mask loss $L_{mask}$ as supervision for the transparency of each pixel.
For some rigid and sharp parts of the face, such as teeth, the pixel-based loss is often difficult to optimize.
Therefore we added the LPIPS loss \cite{zhang2018unreasonable} $L_{lpips}$ based on depth perception. Based on the landmark, we give additional patch L1 loss $L_{lips}$to the lip region \textit{P} to enhance the reconstruction of strongly correlated regions of speech.
Overall, the total loss function is formulated as follows:
\[L = L_{color} + \lambda_1 L_{mask} +\lambda_2 L_{lpips} + \lambda_3 L_{lips} \tag{7}\]

\section{experiments}


\subsection{Experimental Settings}

\subsubsection{Dataset}
We utilize five person-specific videos with audio track as our training material which are also used in \cite{guo2021ad,ye2023geneface} for a fair comparison. Each video lasts for 3 $\sim$ 5 minutes and the frame rate is adjusted to 25 fps. The raw video in AD-NeRF \cite{guo2021ad} is cropped to 450 $\times$ 450, while other videos in \cite{ye2023geneface} are cropped to 512 $\times$ 512. Besides, we also extracted two audio tracks from public demos of SynObama \cite{suwajanakorn2017synthesizing} and NVP \cite{thies2020neural}, named \textbf{Testset A} and \textbf{Testset B} respectively, as part of our evaluation of the sources.
\subsubsection{Implementation Details}
For each video, we randomly initialize 10,000 Gaussian points within a cube of side length 2. Then we optimize our Gaussian properties individually for 10,000 iterations in the static initialization phase and optimize our Gaussian points and deformation field simultaneously for 100,000 iterations. For each 2D hash encoder, we set L = 14, and F = 1, and resolutions increased from 64 to 512. Adam optimizer\cite{kingma2014adam} with a learning rate of 5e-3 for the hash encoder and 1.6e-5 for the deformation network. We perform all of our experiments on a single RTX 3090 GPU.
\subsubsection{Comparing baselines and metrics}
Several typical and popular baseline models are adopted for comparison, including Wav2-Lip \cite{prajwal2020lip}, MakeItTalk \cite{zhou2020makelttalk} and 3D-NeRF based person-specific methods: AD-NeRF \cite{guo2021ad}, RAD-NeRF \cite{tang2022real} and ER-NeRF \cite{li2023efficient}.
The metrics to evaluate talking face models consist of image quality, lip synchronization, and efficiency. 
More specifically, we employ the Peak Signal-to-Noise Ratio (PSNR) and Learned Perceptual Image Patch Similarity (LPIPS) \cite{zhang2018unreasonable} to measure the quality of image generation. To measure lip synchronization, the landmark distance (LMD) \cite{chen2018lip} is adopted to calculate the L2 distance of the mouth region landmarks between the synthesized image and the real picture. We also use SyncNet confidence score (Sync) \cite{chung2017out} which uses a pre-trained SyncNet to measure the lip-sync error between the generated video frames and the corresponding speech segment. Action units error (AUE) \cite{chung2017out} is used to evaluate the degree of activity synchronization of the lip-related facial muscles. Besides, we also compare the time overhead of training and inference between the different methods.
\subsection{Quantitative Evaluation}
\subsubsection{Comparison Settings}
We take two settings for the quantitative comparison: 1) Self-driven setting, the speaker of the audio tracks in the train, and test sets are the same. 2) Cross-driven setting, we use the \textbf{Testset A} and \textbf{Testset B}, whose speakers are unseen during the model training. We train all models on the AD-NeRF \cite{guo2021ad} dataset for fairness. Since the ground truth video is not available for the cross-driven set, Sync and AUE are adopted as evaluation metrics.
\begin{figure}[h]
	\begin{minipage}{0.240\linewidth}
	\vspace{3pt}
	\centerline{\includegraphics[width=\textwidth]{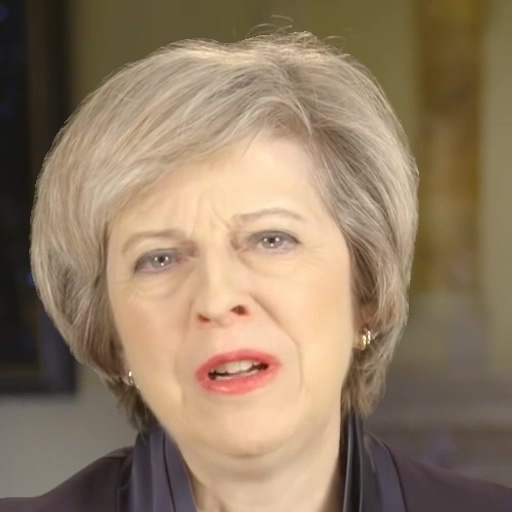}}
	\vspace{3pt}
	\centerline{\includegraphics[width=\textwidth]{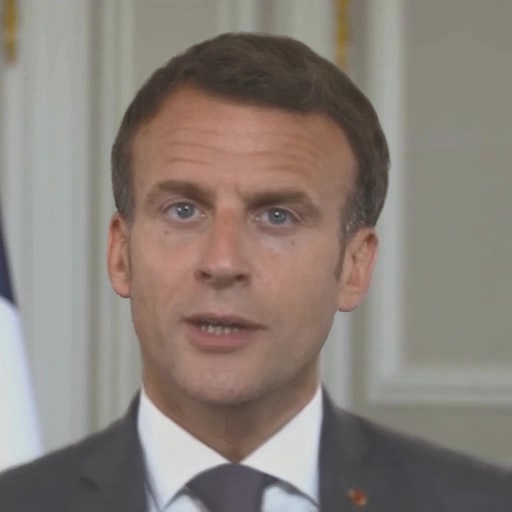}}
	\vspace{3pt}
	\centerline{\includegraphics[width=\textwidth]{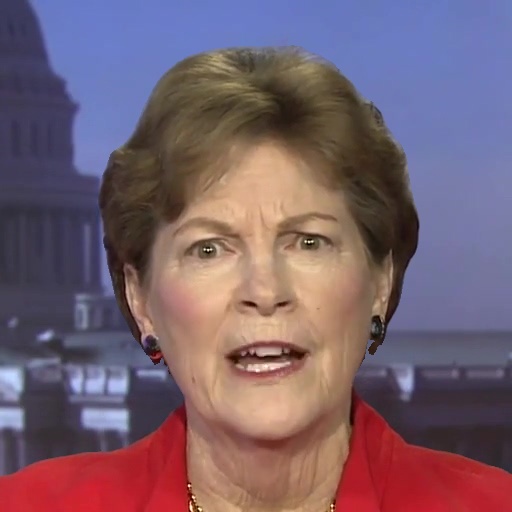}}
	\vspace{3pt}
    \centerline{\includegraphics[width=\textwidth]{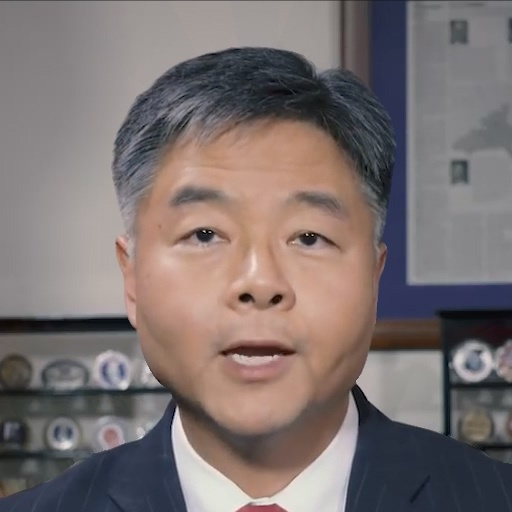}}
	\vspace{3pt}
	\centerline{Ground-Truth}
	\end{minipage}
	\begin{minipage}{0.240\linewidth}
	\vspace{3pt}
	\centerline{\includegraphics[width=\textwidth]{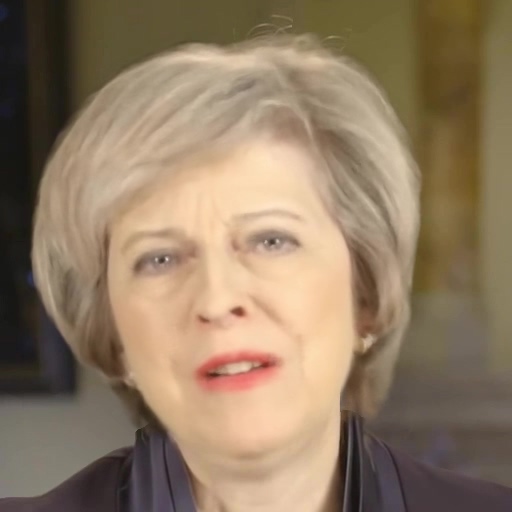}}
	\vspace{3pt}
	\centerline{\includegraphics[width=\textwidth]{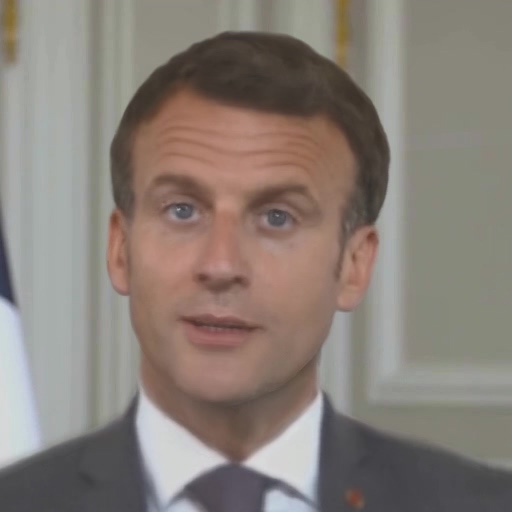}}
	\vspace{3pt}
	\centerline{\includegraphics[width=\textwidth]{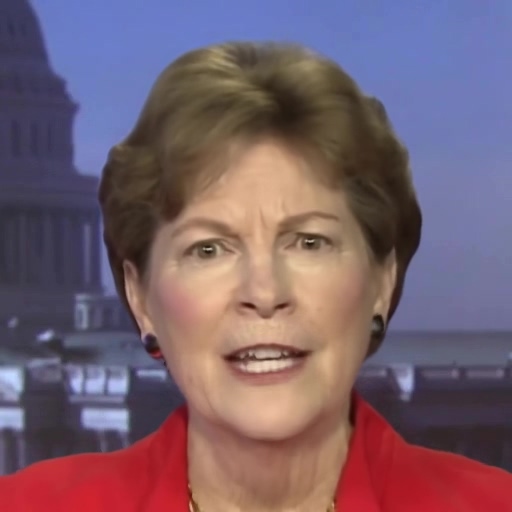}}
	\vspace{3pt}
    \centerline{\includegraphics[width=\textwidth]{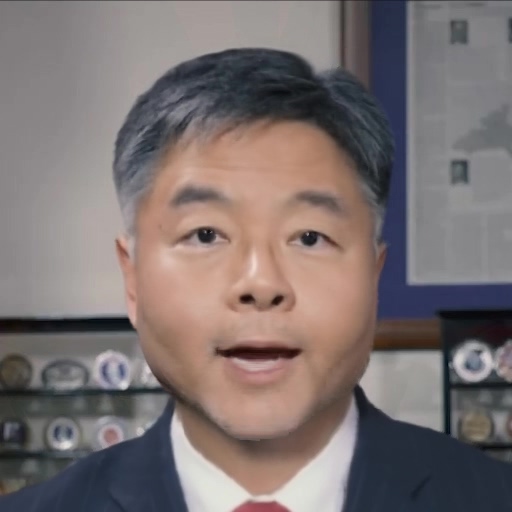}}
	\vspace{3pt}
	\centerline{RAD-NeRF}
\end{minipage}
	\begin{minipage}{0.240\linewidth}
	\vspace{3pt}
	\centerline{\includegraphics[width=\textwidth]{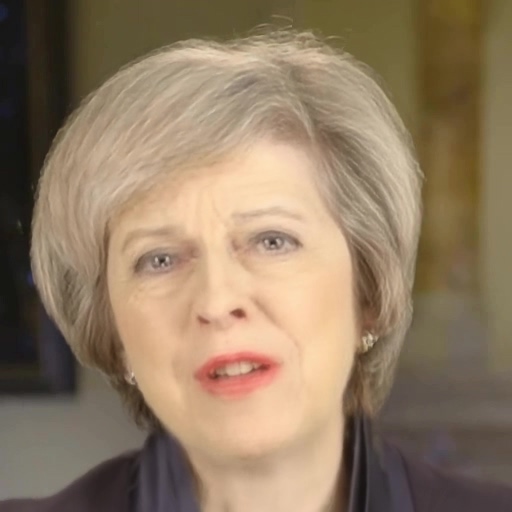}}
    \vspace{3pt}
    \centerline{\includegraphics[width=\textwidth]{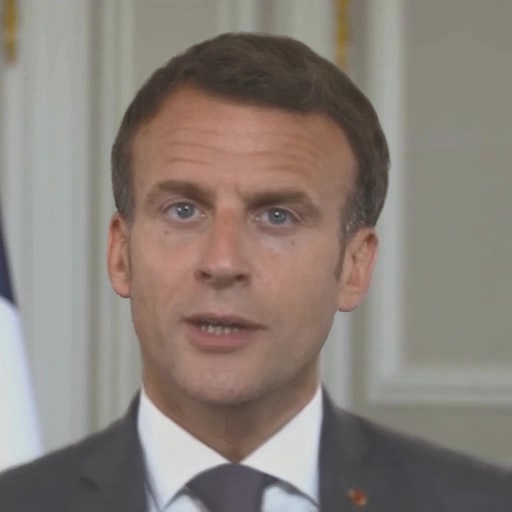}}
    \vspace{3pt}
    \centerline{\includegraphics[width=\textwidth]{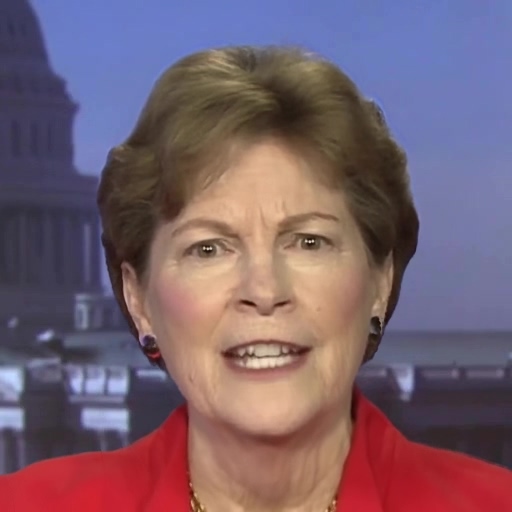}}
    \vspace{3pt}
	\centerline{\includegraphics[width=\textwidth]{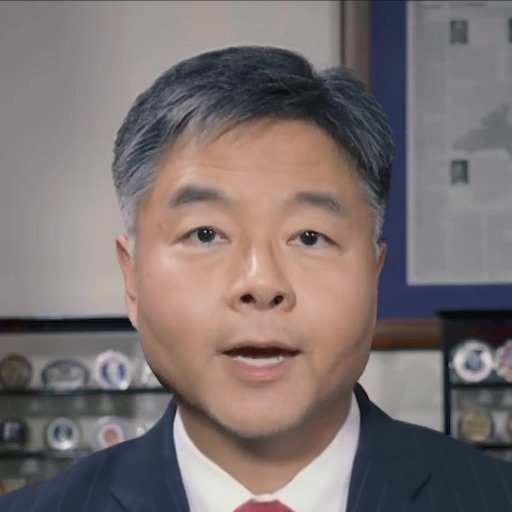}}
	\vspace{3pt}
	\centerline{ER-NeRF}
\end{minipage}
	\begin{minipage}{0.240\linewidth}
	\vspace{3pt}
	\centerline{\includegraphics[width=\textwidth]{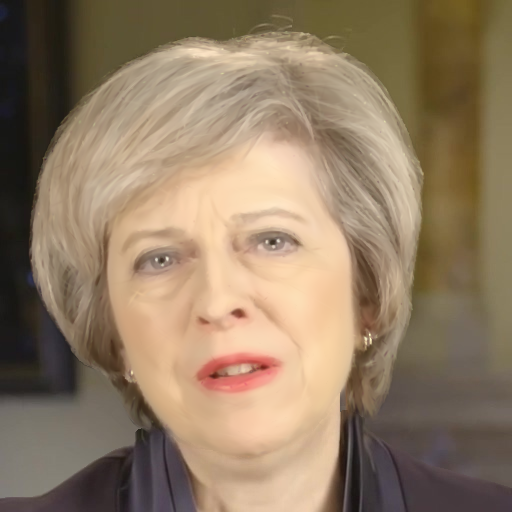}}
	\vspace{3pt}
	\centerline{\includegraphics[width=\textwidth]{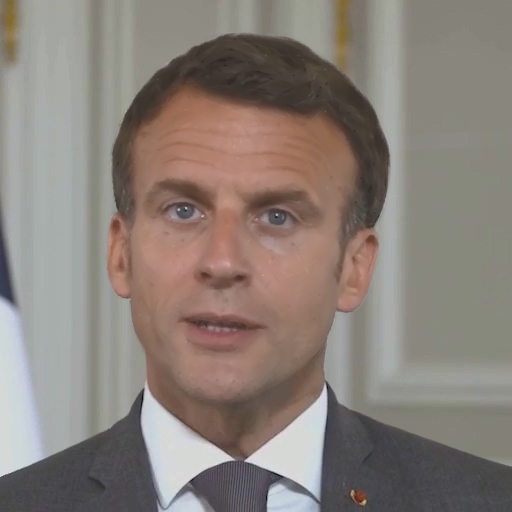}}
	\vspace{3pt}
	\centerline{\includegraphics[width=\textwidth]{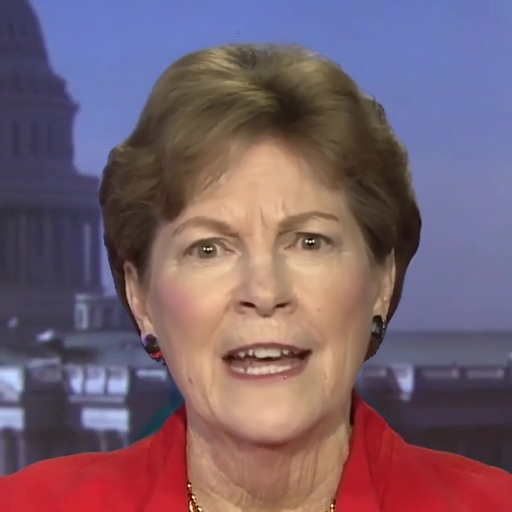}}
	\vspace{3pt}
	\centerline{\includegraphics[width=\textwidth]{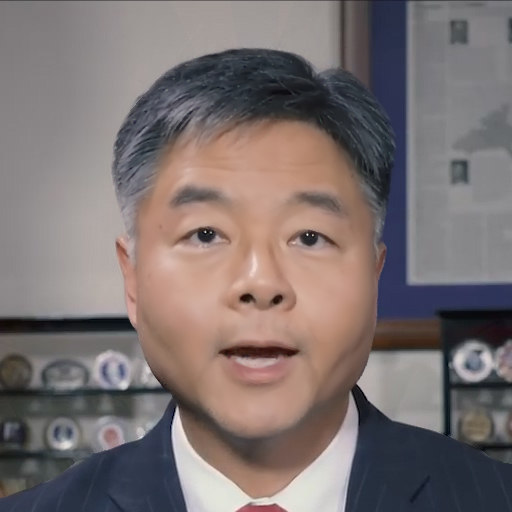}}
	\vspace{3pt}
	\centerline{GSTalker}
\end{minipage}
\caption{\textbf{The qualitative comparison with real-time NeRF-based methods.} We show more generated results compared to the real-time NeRF-based methods. These results include the performance across genders and languages in the self-driven setting.}
\label{f3}
\end{figure}
\begin{table}[h]
\begin{center}
\resizebox{\linewidth}{!}{
\begin{tabular}{lcccc}
   \toprule
   &\multicolumn{2}{c}{TestsetA} &\multicolumn{2}{c}{TestsetB}\\
   Methods &Sync$\uparrow$ &AUE$\downarrow$  &Sync$\uparrow$ &AUE$\downarrow$ \\
   Ground Truth &6.855 &0 &7.486 &0\\
   \hline
   MakeItTalk\cite{zhou2020makelttalk} &5.351 &2.121 &5.091 &1.444\\
   Wav2Lip\cite{prajwal2020lip}  &\textbf{7.912} &1.822 &\textbf{8.149} &1.259\\
   AD-NeRF\cite{guo2021ad}  &4.855 &1.943 &4.698 &1.541\\
   RAD-NeRF\cite{tang2022real} &5.747 &1.788 &6.372 &1.237\\
   ER-NeRF\cite{li2023efficient} &5.830 &\underline{1.782} &\underline{6.800} &\textbf{1.156}\\
   \hline
   Ours &\underline{6.299} &\textbf{1.771} &6.718 &\underline{1.251}\\
  \bottomrule
\end{tabular}}
\end{center}
\bigskip\centering
\caption{\textbf{The quantitative results under the cross-driven setting.} We compare the audio-lips
synchronization and facial motion similarity with the other methods under the cross-driven setting. We show the best and second best results with \textbf{blod} and \underline{underline}, respectively.}
\label{t2}
\end{table}
\subsubsection{Evaluation Results}
We present the quantitative comparison results under the self-driven setting in Table \ref{t1}, Compared with the best competitive baseline model (ER-NeRF), our proposed GSTalker achieves $2\sim 3$ times speedup in both training and inference. Meanwhile, our method presents superior picture quality performance not only at the pixel level but also at the depth perception level. In terms of audio-lip synchronization, our methods achieve the lowest landmark distance on the mouth and promising performance on Sync and AUE. In the cross-driven setting experiment in Table \ref{t2}, our method ranks among the best in both the Sync and AUE metrics, which reveals the excellent generalization ability across the speaker for our proposed method.
\subsection{Qualitative Evaluation}
\subsubsection{Evaluation Results}
We represent the visual results of all methods in the AD-NeRF \cite{guo2021ad} dataset in Figure \ref{f2}. Although Wav2lip and MakeItTalk work with arbitrary portrait inputs and perform well on some metrics, it is still difficult to produce realistic and high-fidelity videos. For the 3D NeRF-based methods, we synthesized both the head and torso parts for Evaluation. AD-NeRF which uses a vanilla NeRF for scene representation, tends to generate ambiguous torso and audio-lip unsynchronized mouth. RAD-NeRF and ER-NeRF, while able to synthesize high-quality torso and facial details, still struggle with fine teeth structure. Further, we compare our method with RAD-NeRF and ER-NeRF on more portraits as presented in Figure \ref{f3}. It is found that our method achieves finer hair and facial details and a clearer torso, as well as sharper teeth and improved audio-lip synchronization.
\subsubsection{User Study}
A user study is employed to further evaluate user-specific real-world experiences. Specifically, we selected 36 generated video results from different methods and surveyed 26 users on the following three aspects: (1) Lip-sync Accuracy; (2) Motion Naturalness; and (3) Image Fidelity. The statistical average results are shown in Table \ref{user study}. Our method yields the best head naturalness performance, in addition to being among the best in terms of lip-sync accuracy and image fidelity. Note that our approach only requires less training time as well as having faster inference speeds
\begin{table}[!h]
\begin{center}
\resizebox{\linewidth}{!}{
\begin{tabular}{lccc}
  \toprule
   Methods &Lip-sync Accuracy &Motion Naturalness &Image Fidelity\\
   \hline
    MakeItTalk &3.55 &3.55 &3.28 \\
    Wav2lip &3.19 &3.26 &2.70 \\
    AD-NeRF &3.93 &3.51 &3.91 \\
    RAD-NeRF &\textbf{4.10} &3.95 &4.16 \\
    ER-NeRF &\underline{4.09} &\underline{4.11} &\textbf{4.22} \\
    \hline
    GSTalke &4.04 &\textbf{4.14} &\underline{4.18} \\
  \bottomrule
\end{tabular}}
\bigskip\centering
\caption{\textbf{User study.} The rating is on a scale of 1-5, the higher the better. We show the best with \textbf{bold} and second best results with \underline{underline}.}\label{user study}
\end{center}
\end{table}
\subsection{Ablation Study}
An ablation study of network architecture and optimization solutions, including the qualitative and quantitative results, are shown in Table \ref{ablation}. The experiments are carried out on the AD-NeRF \cite{guo2021ad} dataset under the self-driven setting.
\begin{table}[h]
\begin{center}
\resizebox{\linewidth}{!}{
\begin{tabular}{lcccc}
  \toprule
   Methods &PSNR$\uparrow$ &LPIPS$\downarrow$ &LMD$\downarrow$ &Sync$\uparrow$\\
   Ground Truth &$\infty$ &0 &0 &7.491\\
   \hline
   w/o static initialization  &34.25 &0.0169 &3.068 &1.506\\
   w/o hash grid  &34.39 &0.0159 &2.798 &5.438\\
   w/o LPIPS loss  &\textbf{34.66} &0.0259 &2.773 &5.305\\
   \hline
   Ours &34.65 &\textbf{0.0151} &\textbf{2.695} &\textbf{5.775}\\
  \bottomrule
\end{tabular}}
\end{center}
\bigskip\centering
\caption{\textbf{Ablation Study} on network architecture and optimization solutions. Comparing image quality and audio-lips synchronization on static Gaussian initialization, multiresolution hash grid, and perceptual loss supervision} 
\label{ablation}
\end{table}
\begin{table}[h]
\begin{center}
\resizebox{\linewidth}{!}{
\begin{tabular}{lcccc}
  \toprule
   Methods &PSNR$\uparrow$ &LPIPS$\downarrow$ &LMD$\downarrow$ &Snyc$\uparrow$\\
   Ground Truth &$\infty$ &0 &0 &7.491\\
   \hline
   Deepspeech\cite{amodei2016deep} &34.65 &0.0151 &2.695 &5.775\\
   Wav2Vec\cite{baevski2020wav2vec}  &34.65 &\textbf{0.0149} &2.661 &5.903\\
   Hubert\cite{hsu2021hubert} &3\textbf{4.66} &0.0152 &\textbf{2.657} &\textbf{5.960}\\
  \bottomrule
\end{tabular}}
\end{center}
\bigskip\centering
\caption{\textbf{Ablation study} on the input audio feature. We compare the effect of different types of speech features on our model and find that speech features \cite{baevski2020wav2vec, hsu2021hubert} trained in a self-supervised manner have better performance.}
\label{table ablation}
\end{table}
\noindent \textbf{Static Gaussian Initialization.}
We employ a static Gaussian initialization stage before training our deformation field to accelerate the convergence of our speech-driven speaker generation. The results in Table \ref{ablation} show that although the comparable image quality can still be achieved without static initialization, it is difficult to synthesize audio-lip synchronized videos, which suggests that a static Gaussian initialization stage can greatly accelerate the convergence of our audio-driven deformation field and achieve higher audio-lip synchronization accuracy. 

\noindent \textbf{Muti-resolution hash grid.}
The role of the multi-resolution hash tri-plane encoder in our model is to obtain more compact spatial features. In the ablation study experiment, we directly feed the 3D coordinates of Gaussians into the deformation network and find that it degrades both the image quality and Lip-sync. This indicates that the multi-resolution hash grid can better learn the motion correlation between Gaussians.

\noindent \textbf{Perceptual loss.}
We also conduct experiments in the case of not adding LPIPS loss only using pixel-level loss as a supervisor. In the case of a tiny difference in PSNR, we can visually find it difficult to synthesize sharp tooth details, which shows that LPIPS loss can synthesize better visual effects.

\section{CONCLUSION}
In this paper, we present a novel real-time audio-driven talking face generation framework that employs a Gaussian deformation field conditioned on speech and poses to learn the motion of the face and torso, respectively. Our method achieves high quality while significantly improving training and inference efficiency compared with 2D-based and 3D NeRF-based methods. Extensive experiments demonstrate the effectiveness of our proposed methods.

\bibliographystyle{ACM-Reference-Format}
\bibliography{main}
\end{document}